\documentclass{article}
\usepackage{spconf,amsmath,epsfig,multirow,graphicx,amssymb}
\usepackage[switch]{lineno}
\usepackage[noadjust]{cite}

\let\OLDthebibliography\thebibliography
\renewcommand\thebibliography[1]{
  \OLDthebibliography{#1}
  \setlength{\parskip}{0pt}
  \setlength{\itemsep}{0pt plus 0.3ex}
}

\begin{document}\sloppy

\def\x{{\mathbf x}}
\def\L{{\cal L}}

\title{Exploring Linear Feature Disentanglement For Neural Networks}
%
\name{Tiantian He$^{1}$, Zhibin Li$^{2}$, Yongshun Gong$^{1*}$, Yazhou Yao$^{3}$, Xiushan Nie$^{4}$, Yilong Yin$^{1*}$\thanks{*Corresponding authors.}}
\address{$^{1}$Shandong University, $^{2}$CSIRO, $^{3}$Nanjing University of Science and Technology\\$^{4}$ Shandong Jianzhu University}

\maketitle

\begin{abstract}
Non-linear activation functions, e.g., Sigmoid, ReLU, and Tanh, have achieved great success in neural networks (NNs). Due to the complex non-linear characteristic of samples, the objective of those activation functions is to project samples from their original feature space to a linear separable feature space. This phenomenon ignites our interest in exploring whether all features need to be transformed by all non-linear functions in current typical NNs, i.e., whether there exists a part of features arriving at the linear separable feature space in the intermediate layers, that does not require further non-linear variation but an affine transformation instead. To validate the above hypothesis, we explore the problem of linear feature disentanglement for neural networks in this paper. Specifically, we devise a learnable mask module to distinguish between linear and non-linear features. Through our designed experiments we found that some features reach the linearly separable space earlier than  the others and can be detached partly from the NNs. The explored method also provides a readily feasible pruning strategy which barely affects the performance of the  original model. We conduct our experiments on four datasets and present promising results. 
\end{abstract}
\section{Introduction}
\label{sec:intro}

Neutral networks (NNs) have achieved great success in a large variety of fields, e.g., multi-media data processing, natural language processing, and speech recognition~\cite{lecun2015deep}. Non-linear activation functions, e.g., Sigmoid, ReLU~\cite{2010Rectified}, and Tanh, play important roles in NNs. The accumulation of multiple activation functions enables the NN to fit any continuous function, as well as projects samples with complex non-linear characteristics from their original feature space to a linear separable feature space~\cite{hornik1989multilayer}. 

\begin{figure}[t]
\begin{minipage}[b]{0.48\linewidth}
  \centering
  \includegraphics[width=1.0\textwidth]{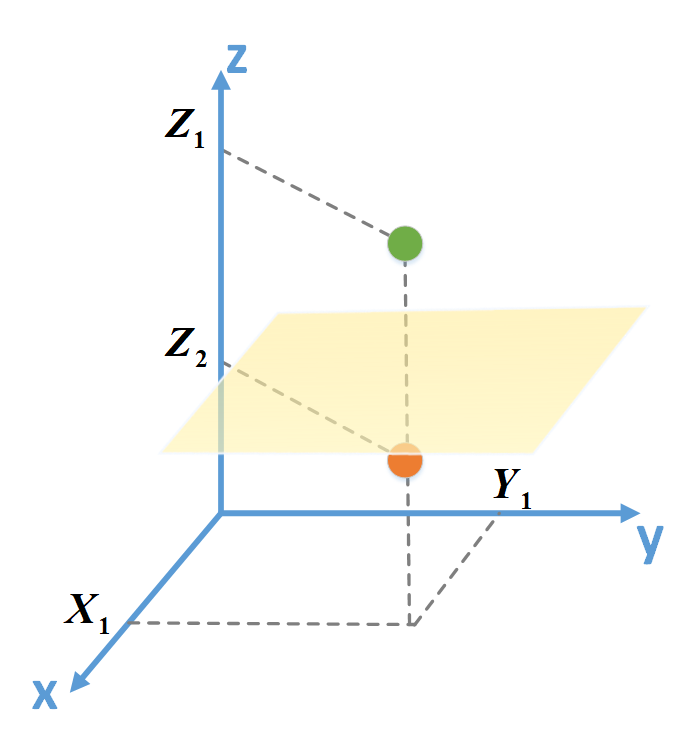}
  \centerline{(a)}\medskip
\end{minipage}
\begin{minipage}[b]{0.5\linewidth}
  \centering
  \includegraphics[width=0.74\textwidth]{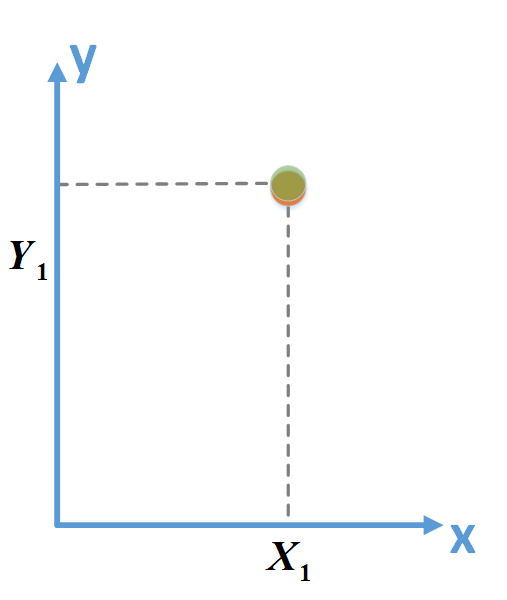}
  \centerline{(b)}\medskip
\end{minipage}
\caption{Illustration of two samples.}
\label{fig:resa}
\end{figure}

In this study, we aim to explore that whether all features need to be transformed by non-linear activation functions in current typical NNs, i.e., whether there exists a part of features have reached a linearly separable space earlier than the others during the learning process. Accordingly, such features do not need to pass the non-linear transformation while using an affine transformation instead. As illustrated in (a) of Fig.~\ref{fig:resa}, two samples are presented into three dimensions. It is apparent that these two samples can be distinguished by their features in z axis via a linear interface, but linearly indistinguishable in the x and y axes because they are projected to the same point (see (b) in Fig.~\ref{fig:resa}). In this case, we illustrate that the features in the third dimension (z) are linear separable, and features in the first two dimensions (x and y) are required further non-linear variation. Note that, these linear distinguishable features do not mean that samples can be classified based solely on them, it is a complex learning process.

To verify our hypothesis, we explore the problem of linear feature disentanglement in neural networks. Specifically, we design a learnable mask module to distinguish between linear features and non-linear features. The mask module generates two complementary binary masks for disentangling linear features. To overcome the non-differentiable problem of the binary mask in the back-propagation, we adopt the straight-through estimator ~\cite{2013Estimating} to estimate the gradient. From the results of experiments, we discovered some phenomena as we expect: there exist the linear features which appear in the intermediate layers, and the proportions of them can converge at the end of the training process. Moreover, our method barely deteriorates the model performance. 

Furthermore, we also explore the potential of the proposed network structure in network pruning. The intuition
behind this design is to leverage the linear property of features to build a fast track, where linear features can be connected to the last layer of neural networks directly. The main contributions of this paper are summarized as follows:


\begin{itemize}
\item We provide the first attempt on distinguishing linear features during the learning process of NNs. A module using learnable masks is proposed according to disentangle linear features.
\item We include the binarization of the vector generated by the mask module into the training, and use the straight-through estimator~\cite{2013Estimating} to alleviate the non-differentiable problem of the binarization function.
\item We conduct extensive experiments on convolutional neural network (CNN). Experimental results confirm that a part of features reached the linearly separable space earlier than others. Accordingly using our disentanglement strategy will barely affect the model performance. 
\item We apply the proposed method in network pruning and show some promising results.   
\end{itemize}

\section{Related work}

In this section, we briefly review some related works on feature transformation in NNs. In the early studies, the Logistic Sigmoid and Tanh have been widely used in the traditional NNs. However, they suffer from vanishing gradient especially when models go deeper. Hinton et al.~\cite{2010Rectified} propose the Rectified Linear Unit (ReLU), which makes the training of deep NNs quicker and overcomes the problem of vanishing gradient. It is still one of the most popular activation functions and is extensively applied in deep models due to its simplicity, generality and effectiveness. Subsequently, various variants of ReLU have emerged, e.g., LeakyReLU~\cite{maas2013rectifier}, MTLU~\cite{MTLU}. These variants make up for the dead neuron issue of ReLU. Besides, some adaptive activation functions have been investigated recently. They contain some trainable parameters tuning the non-linearity according to the data. For example, Ramachandran et al.~\cite{swish} propose an adaptive function called Swish. It introduces a learnable parameter which controls the degree of interpolation between the linear function and the ReLU function. The recently proposed general activation function, ACON~\cite{ma2021activate} is in fact an extension of Swish, whose switch factor allows every neuron can be activated with different degrees. With the expansion of the non-linear search space, these trainable activation functions also increase the time and space complexity. 

In prior applications of activation functions, all features are fed to the non-linear activation functions. To our best knowledge, there is limited work on investigating whether it is necessary to pass all of features through non-linear activation functions. This paper provides the first attempt on the linear feature disentanglement problem during the NN learning process.

\section{Proposed Method}

In this section, we provide a feasible scheme to identify the non-linear and linear features. We present the linear feature disentanglement for CNN models in the paper.

\begin{figure}[t]
\centering
\includegraphics[width=0.45\textwidth]{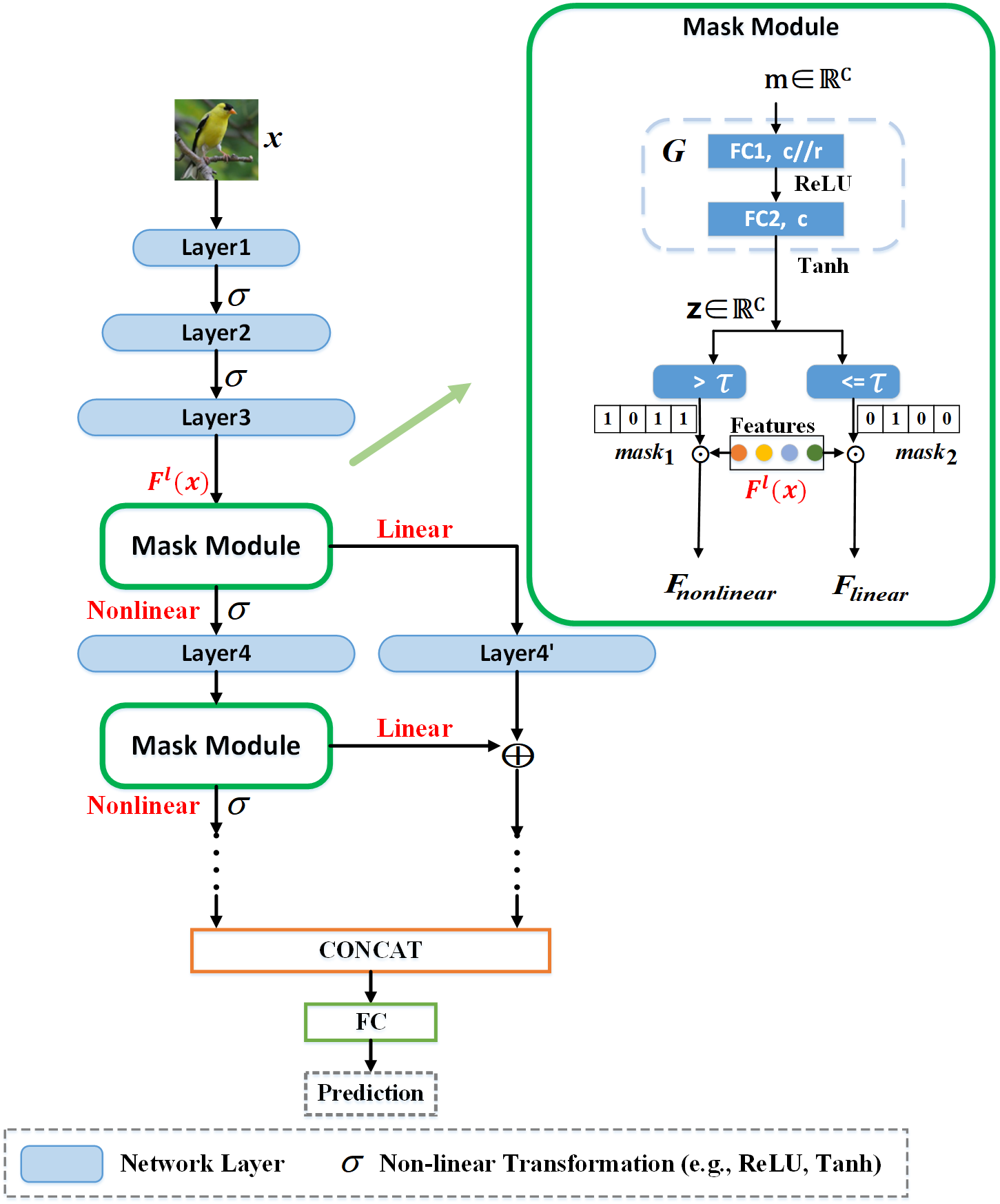}
\caption{Overall pipeline of our method. In the mask module, we set the threshold $\tau$ to 0 in our implementation.}
\label{fig:res1}
\end{figure}

\textbf{Overall Structure.} The overall structure  of our method is illustrated in Fig.~\ref{fig:res1}. The mask module in it generates two complementary binary masks for selecting the linear and non-linear features from the output features of a layer. The mask modules divide the features into linear and non-linear part. While the linear features no longer undergo any non-linear transformation because they have been mapped into the linearly separable space well, and just need some appropriate affine transformations. Finally, the two types of features are concatenated as the final features as the input of the final decision layer.

\textbf{Mask Module.} Specifically, the mask module as shown in the green box of Figure~\ref{fig:res1} learns a function as below:
\begin{eqnarray}
\mathbf{z} =Tanh(G(\mathbf{m})) \in \mathbb{R}^c.
\end{eqnarray}
Here, $\mathbf{m}\in \mathbb{R}^c$ is a randomly initialized vector, where $c$ is the channel number of the input features for the mask module in CNN. $G$ is a function defined as $G$ : $\mathbf{m}\rightarrow \mathbf{g}, \mathbf{g}\in \mathbb{R}^c$. It consists of two fully-connected layers, $\mathbf{g} = FC_2(ReLU(FC_1(\mathbf{m})))$, where $FC_1$ and $FC_2$ are two linear transformations,  ReLU~\cite{2010Rectified} is the non-linear activation function. And there is another non-linear activation function Tanh, which normalizes the output of the function $G$ to [-1, 1] for stabilizing the training process of the whole network.

Notice that $\mathbf{z}$ is a real-valued vector, we need to transform it to a binary vector to indicate which feature is linearly separable. The final two masks are expressed as below:
\begin{eqnarray}
	\mathbf{mask_1}[i] = \begin{cases}
	1, &\mathbf{z}[i] > 0\\
	0, &\mathbf{z}[i] \leq 0	,i \in [1,c],
		   \end{cases}
\end{eqnarray}
\begin{eqnarray}
	\mathbf{mask_2} = \mathbf{1}-\mathbf{mask_1},
\end{eqnarray}
where $\mathbf{mask_1}$ and $\mathbf{mask_2}$ correspond to the non-linear features and linear features respectively. The purpose of $\mathbf{mask_1}$ and $\mathbf{mask_2}$  is to distinguish which channels (or convolution kernels) in CNN models can be recognized as the linear channels. Given a input image $\mathbf{x}$, we define the non-linear feature and linear feature as: $F_{nonlinear}(\mathbf{x}) = \mathbf{mask_1}\odot F^{l}(\mathbf{x})$ and $F_{linear}(\mathbf{x}) = \mathbf{mask_2}\odot F^{l}(\mathbf{x})$, where $F^{l}(\mathbf{x})$ denotes the output feature of the $l$-th convolutional layer and $\odot$ denotes that each element of $\mathbf{mask_1}$ (or $\mathbf{mask_2}$) is multiplied by the corresponding channel of feature $F_{nonlinear}(\mathbf{x})$ (or $F_{linear}(\mathbf{x})$). And the non-zero channels in $F_{linear}(\mathbf{x})$ correspond exactly to the linear kernels in the $l$-th convolutional layer, while the rest correspond to the non-linear kernels. Obviously, Eq.(2) and Eq.(3) are non-differentiable. In this paper, we leverage the method of straight-through estimator~\cite{2013Estimating} and define the gradients of both  $\mathbf{mask_1}$ w.r.t $\mathbf{z}$ and $\mathbf{mask_2}$ w.r.t $\mathbf{z}$ to $\mathbf{1}$. So the well-defined $\frac{\partial F_{nonlinear}(\mathbf{x})}{\partial \mathbf{mask_1}}$ and $\frac{\partial F_{linear}(\mathbf{x})}{\partial \mathbf{mask_2}}$ can be used as approximations for $\frac{\partial F_{nonlinear}(\mathbf{x})}{\partial \mathbf{z}}$ and $\frac{\partial F_{linear}(\mathbf{x})}{\partial \mathbf{z}}$, respectively. 

We set every element in $\mathbf{z}$ $>$ 0 in the network initialization, because we suppose all features need to undergo non-linear transformations at first and then observe changes in the proportion of non-linear and linear features. The initialization  conduces to optimize better the network at the beginning of the training process. And as the training phase progresses, the mask modules will be able to map the corresponding features well to the linearly separable space.  
  
The mask module is a relatively straightforward component, so it is easy to be plugged into various CNN architectures, as well as MLPs and RNNs without special design. And the whole network can be trained in an end-to-end manner via back-propagation thanks to the straight-through estimator.

\begin{figure}[t]
\begin{minipage}[b]{0.48\linewidth}
  \centering
  \centerline{\epsfig{figure=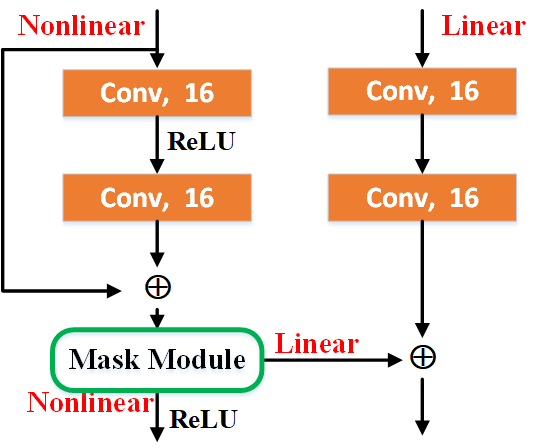,width=4.0cm}}
  \centerline{(a)}\medskip
\end{minipage}
\begin{minipage}[b]{0.48\linewidth}
  \centering
  \centerline{\epsfig{figure=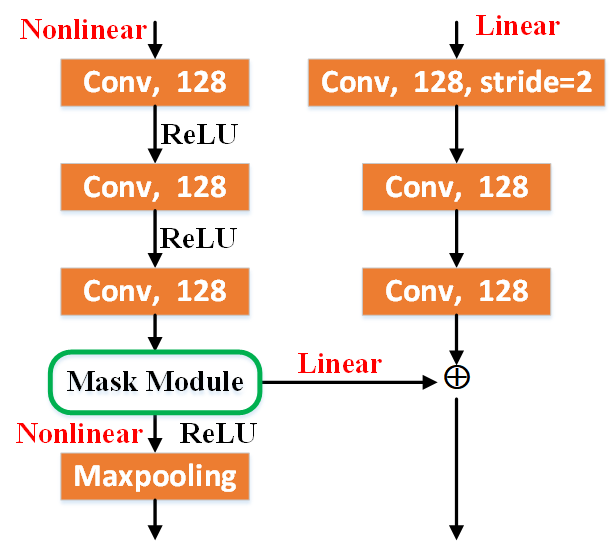,width=4.0cm}}
  \centerline{(b)}\medskip
\end{minipage}
\caption{(a) The schema of the residual module in ResNet-56\_M. (b) The convolutional block in VGG-16\_M.}
\label{fig:res3}
\end{figure}

\section{Experiments}

In this section, the proposed method is empirically investigated on seven common benchmark datasets relevant to classification tasks, CIFAR-10, CIFAR-100~\cite{krizhevsky2009learning}, SVHN~\cite{2011Reading}, STL-10~\cite{2011An}. And we study the performance of our method on two typical CNN models (ResNet-56~\cite{he2016deep} and VGG-16 described in ~\cite{li2016pruning}) with the above four datasets. Finally, we combine our method with network pruning and conduct relevant exploratory experiments. 

\subsection{Experiments setting}

\textbf{Datasets.} CIFAR-10 and CIFAR-100 consist of 50k training images and 10k testing images in 10 and 100 classes respectively. For these two datasets, we hold out 5k images from training images for validation. The SVHN dataset has 73,257 images for training and 26,032 images for testing from 10 categories. In addition, there are 531,131 additional images in SVHN, but we did not use the images due to restricted computational resources. STL-10 contains ${96}\times{96}$ natural images belonging to 10 classes. And most of the images in the dataset are unlabeled, so we use only the labeled images. There are 5k training and 8k testing labeled images. 

\textbf{Architecture setting.} We use ResNet-56 and VGG-16 which are widely used for image classification as the backbone networks. Fig.~\ref{fig:res3} depicts the structures of ResNet-56\_M and VGG-16\_M. Indeed, we add a mask module before the non-linear activation function at the end of each residual block on ResNet-56. while on VGG-16, we do the same before each max-pooling layer. According to the actual situation, the number and location of mask modules can be changed flexibly. Note that, ResNet-56 has three stages of residual block, so we add the mask modules to all three stages on ResNet-56\_M\_A, last two stages on ResNet-56\_M\_B and last stage on ResNet-56\_M\_D. And VGG-16\_M has five mask modules.

\textbf{Training setting.} To get the baseline accuracies for ResNet-56 and VGG-16, we follow the same training schedule as ~\cite{he2016deep}. For a fair comparison, we train our models with mask modules utilizing the same training scheme as baseline models, except that we optimize the mask modules via Adam with an initial learning rate of 0.001 and a weight decay of 0.0001.  

\begin{figure}[t]
\begin{minipage}[b]{0.48\linewidth}
  \centering
  \centerline{\epsfig{figure=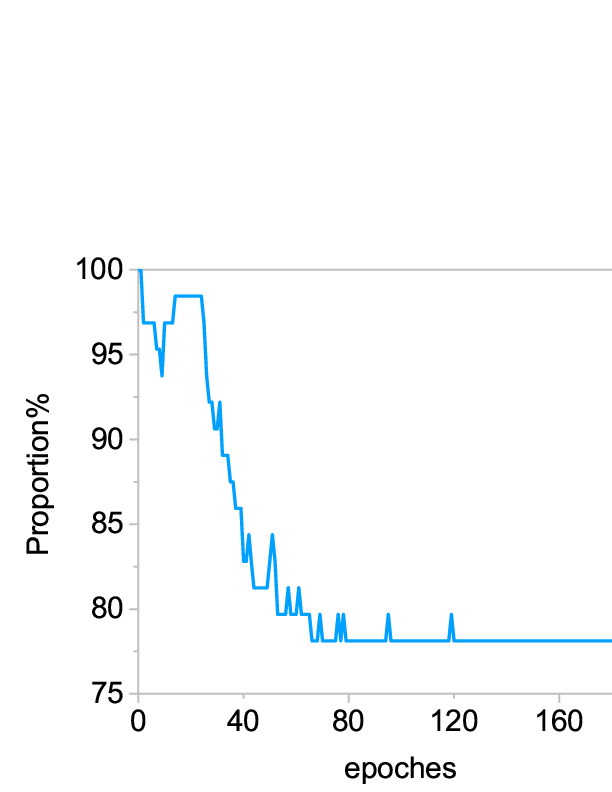,width=4.0cm}}
  \centerline{(a)}\medskip
\end{minipage}
\begin{minipage}[b]{0.48\linewidth}
  \centering
  \centerline{\epsfig{figure=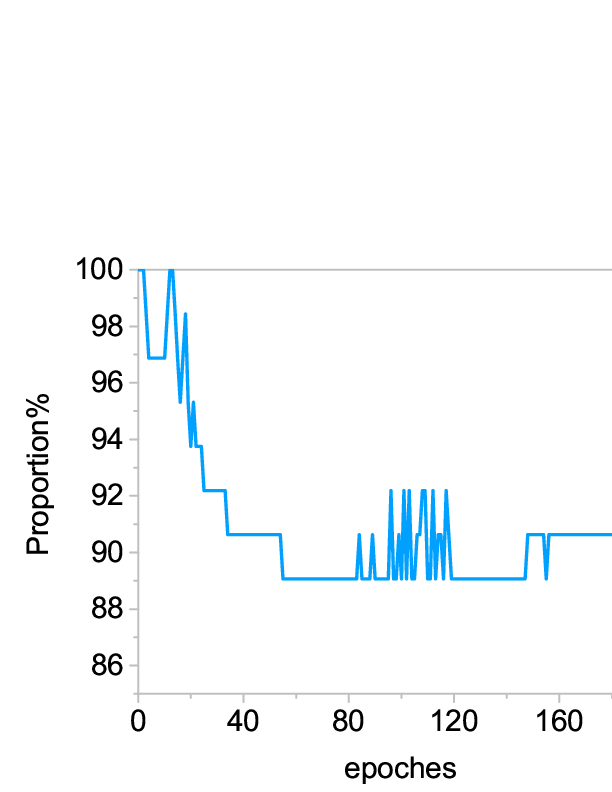,width=4.0cm}}
  \centerline{(b)}\medskip
\end{minipage}
\begin{minipage}[b]{.48\linewidth}
  \centering
  \centerline{\epsfig{figure=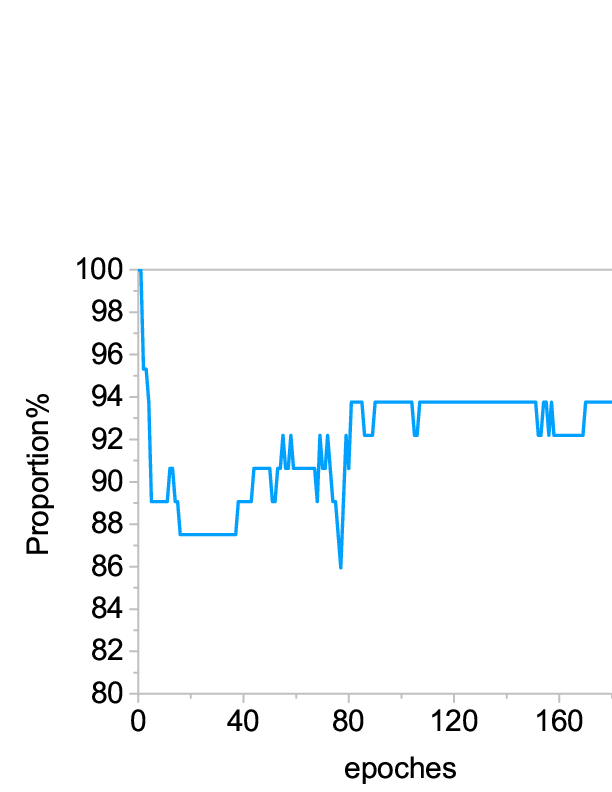,width=4.0cm}}
  \centerline{(c)}\medskip
\end{minipage}
\hfill
\begin{minipage}[b]{0.48\linewidth}
  \centering
  \centerline{\epsfig{figure=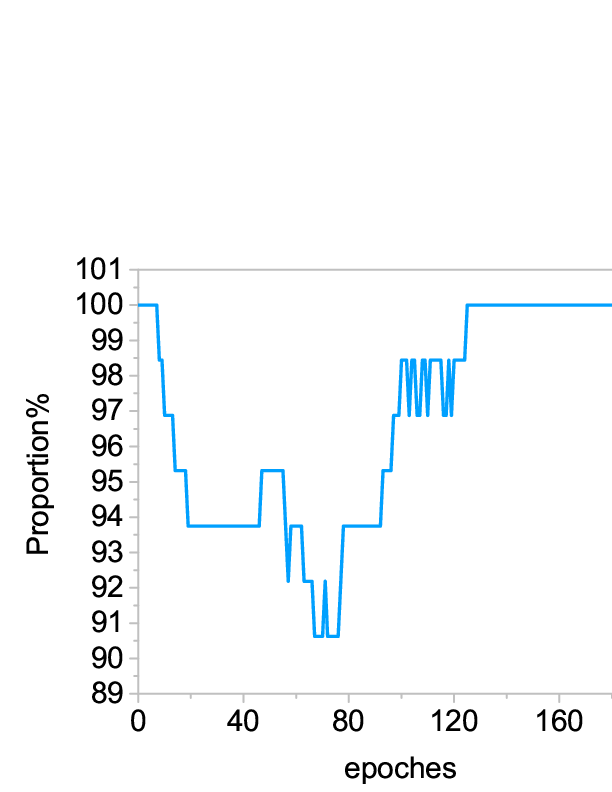,width=4.0cm}}
  \centerline{(d)}\medskip
\end{minipage}
\caption{Proportion curves of some non-linear features for ResNet-56\_M\_D on CIFAR-10. The four subfigures above show the results of non-linear features extracted by the first two and the last two mask modules.}
\label{fig:rescnn}
\end{figure}

\begin{figure}[t]
\begin{minipage}[b]{0.48\linewidth}
  \centering
  \centerline{\epsfig{figure=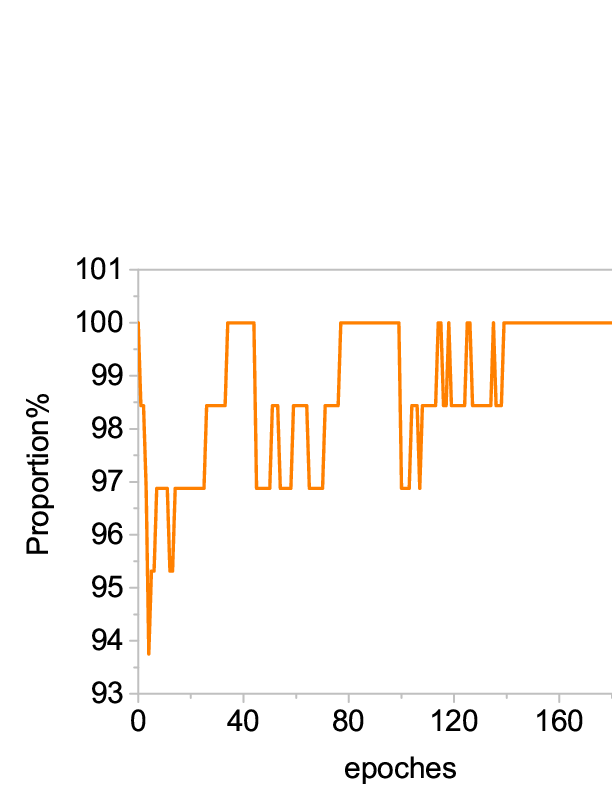,width=4.0cm}}
  \centerline{(a)}\medskip
\end{minipage}
\begin{minipage}[b]{0.48\linewidth}
  \centering
  \centerline{\epsfig{figure=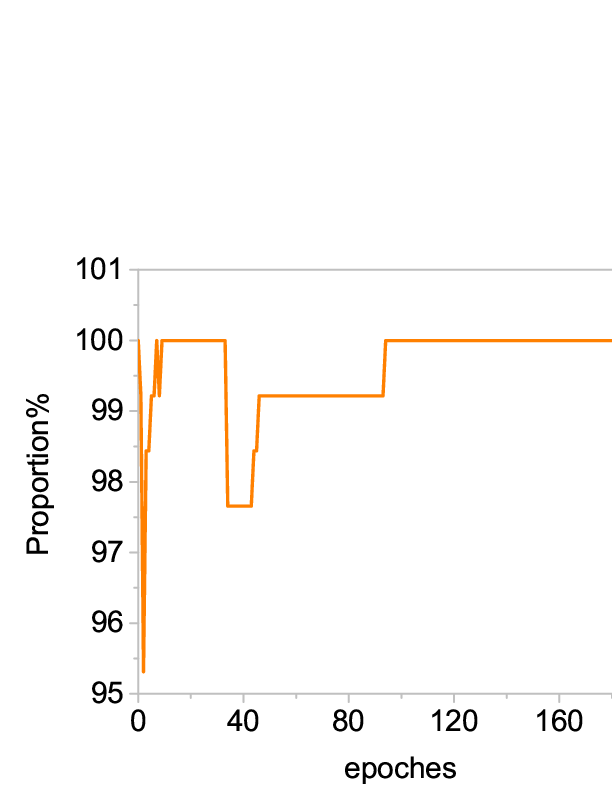,width=4.0cm}}
  \centerline{(b)}\medskip
\end{minipage}
\begin{minipage}[b]{.48\linewidth}
  \centering
  \centerline{\epsfig{figure=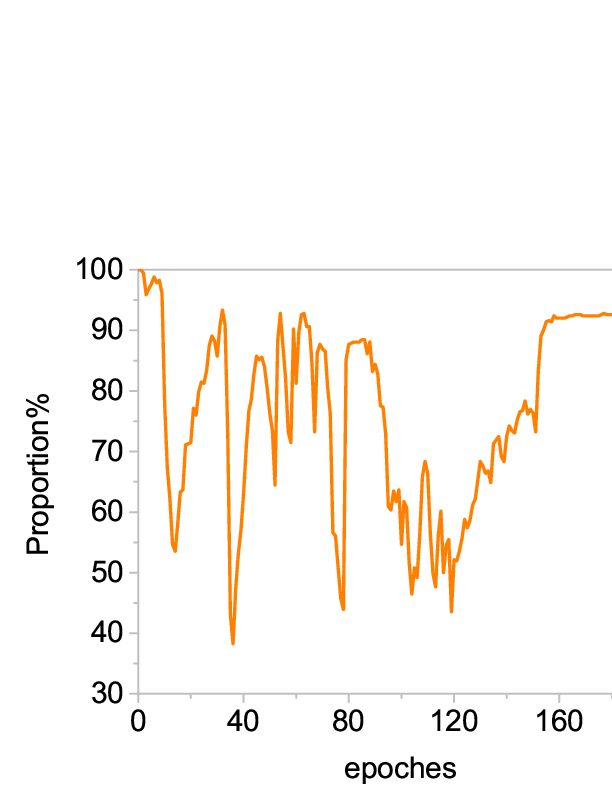,width=4.0cm}}
  \centerline{(c)}\medskip
\end{minipage}
\hfill
\begin{minipage}[b]{0.48\linewidth}
  \centering
  \centerline{\epsfig{figure=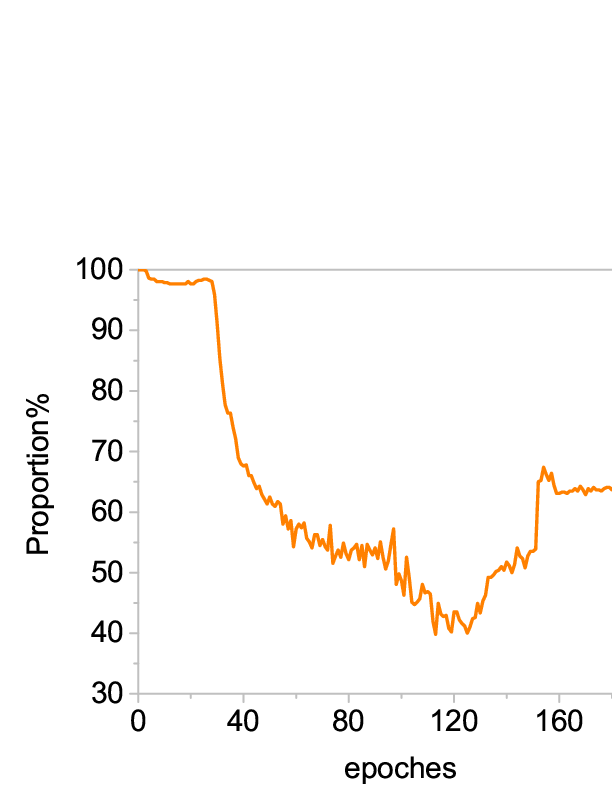,width=4.0cm}}
  \centerline{(d)}\medskip
\end{minipage}
\caption{Proportion curves of some non-linear features for VGG-16\_M on CIFAR-10. The four subfigures above show the results of non-linear features extracted by the first two and the last two mask modules.}
\label{fig:rescnnVGG}
\end{figure}

\subsection{Results and Analysis}

The purpose of the experiments is finding a part of the features that have been mapped to the linearly separable space early. And then we can verify the rationality of our method by top-1 accuracy and the trend of the proportion of features requiring non-linear activation functions. The proportion of the non-linear features in the $l$-th mask module could be formulated as $P_{nonlinear} = \frac{\sum_{i=1}^{c}\mathbf{mask_{1}}^{l}(i)}{c}$ , where $\mathbf{mask_{1}}^{l}(i)$ denotes the $i$-th element of the mask corresponding to the non-linear features in the $l$-th layer.

\begin{table}[hb]
\centering
\caption{Classification accuracy on the CIFAR-10/100, SVHN and STL-10 test sets. The best results are in bold.}\label{tab:1}
\scalebox{0.9}{
\begin{tabular}{c|c|c|c|c}
\hline
Model           & CIFAR-10 & CIFAR-100 & SVHN  & STL-10  \\ \hline
ResNet-56(base)       & 93.07    & \textbf{70.73}     & 96.66 & 76.70   \\
ResNet-56\_M\_A & 93.14    & 70.06     & 96.55 & \textbf{77.01}   \\
ResNet-56\_M\_B & 92.22    & 70.25     & \textbf{96.80} & 75.71   \\
ResNet-56\_M\_D & \textbf{93.28}    & 70.65     & 96.57 & 76.79   \\ \hline
VGG-16(base)          & 93.04    & \textbf{72.54}     & \textbf{96.53} & \textbf{80.64}   \\
VGG-16\_M       & \textbf{93.40}    & 71.92     & 96.41 & 79.69  \\ \hline
\end{tabular}
}
\end{table}

\begin{figure*}[ht]
\centering
\includegraphics[width=1\textwidth]{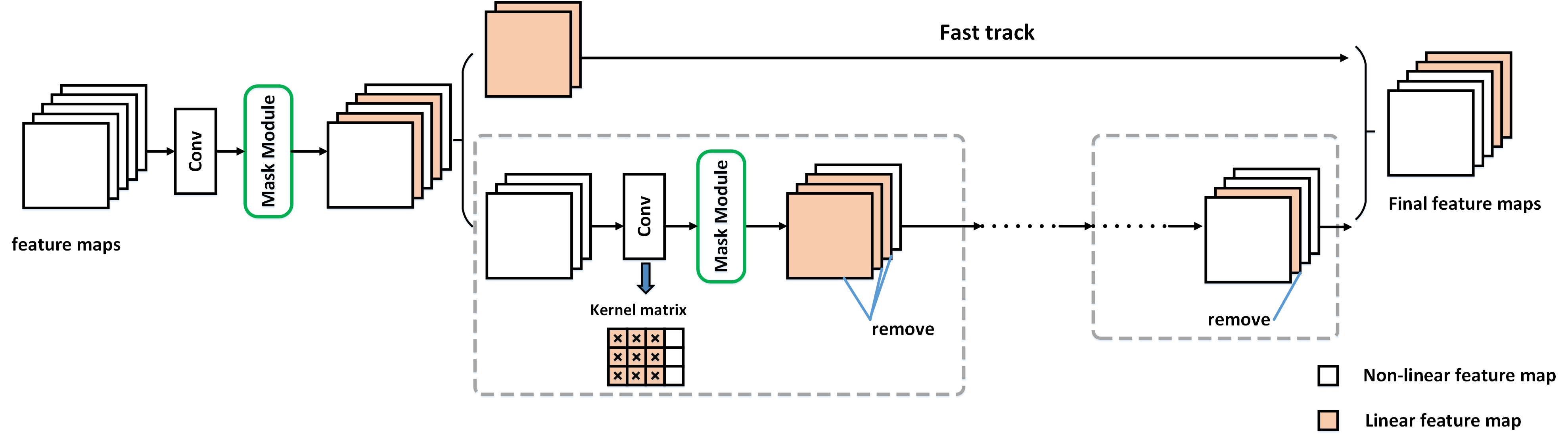}
\caption{Removing linear feature maps results in pruning of corresponding filters.}
\label{fig:resprune}
\end{figure*}

We analyse the performance on CIFAR-10/100, SVHN, STL-10, comparing against two popular CNNs, including ResNet-56 and VGG-16. Firstly, as shown in Fig.~\ref{fig:rescnn}, the proportions of non-linear features have converged eventually in all layers. The phenomenon proves that some intermediate layers in the network already have the ability to map the corresponding features to the linearly separable space. And in terms of the change curves from different layers, we find that if the proportion of non-linear features extracted in the first few layers are  low, the proportion in the last few layers tends to be higher. After linear features are extracted in the first few layers, in order to maintain the performance of the network, more non-linear features need non-linear transformations. On the other hand, we observe the contrary phenomenon from Fig.~\ref{fig:rescnnVGG}. When few or even no linear features are extracted, the representation power of layers closer to the output are relatively stronger during training. So the proportion of non-linear features of last few layers can decrease and finally converge. The difference between ResNet-56\_M\_D and VGG-16\_M might be caused by the depth of models. In particular,  ResNet-56\_M\_D is deeper then VGG-16\_M. Moreover, the mask modules of ResNet-56\_M\_D are equipped in the last stage, and the network parameters in the stage have greater ability to map corresponding features to the linear separable space. So the first mask module of ResNet-56\_M\_D can extract more than 20\% linear features at last. These observations firmly support our proposed assumption.

The experiment results are reported in Table~\ref{tab:1}. For the CIFAR-10 dataset, it is evident that our models ResNet-56\_M\_D and VGG-16\_M are performing better than the baseline models. It seems that the number and location of mask modules can affect the final performance. And for the CIFAR-100 dataset, the performance of our models is somewhat inferior to the baseline models, but the gap between them is marginal. For the SVHN and STL-10 datasets, our models based on ResNet-56 (ResNet-56\_M\_B on SVHN and ResNet-56\_M\_A on STL-10) outperform the baseline models, while two models with VGG-16\_M achieve worse performance than original VGG-16, especially on SVHN. From the results of experiments, our method is not perfect in terms of accuracy. Nonetheless, we provide some insights on the evolution of linear/non-linear features and the disentanglement problem.

\subsection{Exploration on Network Pruning}

Considering the linear features that have been mapped to the linearly separable space, we could set up a fast track to export the linear features which are disentangled by the first mask module to the last decision layer of the network. As for the linear features and non-linear features extracted from subsequent layers, we use only non-linear features, while ignore the linear features. That is, we leave out the network parameters associated with the linear features, except for the linear features extracted from the first mask module (see Fig.~\ref{fig:resprune}). We conduct the experiments on CIFAR-10 and SVHN with our pre-trained networks: ResNet-56\_M and VGG-16\_M.  The masks generated by the mask modules in above two models are made use of to guide pruning. For retraining, we fine-tune the networks for 40 epochs after pruning, with the learning rate of 0.001 and divided by 10 at epochs 10 and 20.

We compare our pruning algorithm with some existing frameworks, e.g., L1~\cite{li2016pruning}, KSE~\cite{KSE}, SFP~\cite{2018Soft}, GAL~\cite{2019Towards}, HRank~\cite{lin2020hrank}, Variational~\cite{zhao2019variational}, FPGM~\cite{He_2019_CVPR}. Table~\ref{tab:4} shows the experimental results. The top-1 accuracy drop between pruned model and the baseline model and the reduction ratios of parameters are reported.

\textbf{CIFAR-10.} For the CIFAR-10 dataset, we prune the network based on pre-trained ResNet-56\_M\_D and VGG-16\_M. As shown in Table~\ref{tab:4}, we observe a gap between our method and others in parameters reduction. But it is noteworthy that the accuracies of our method significantly increase with parameters reduction compared to the original ResNet-56 and VGG-16. The results suggest a great potential  of the feature disentanglement.  

\textbf{SVHN.} For the SVHN dataset, we prune the network based on pre-trained ResNet-56\_M\_B. Compared with original ResNet-56, our method achieves 18.7\% parameters reduction with a 0.21\% accuracy increase. Significantly, our accuracy drop is the smallest, and the reduction of parameters is more than the classic channel pruning method, L1~\cite{li2016pruning}. The results show that our pruning algorithm achieves competitive performance.

\begin{table}[t]
\centering
\caption{Comparison of the pruned ResNet-56 and VGG-16 on CIFAR-10 and SVHN. The "Acc.$\downarrow$" is the accuracy drop between pruned model and the baseline model, the smaller, the better. The best results are in bold and the second best are underlined. } \label{tab:4}
\scalebox{0.9}
{
\begin{tabular}{c|c|c|c}
\hline
Model                                                                         & Method & Acc.$\downarrow$\% & Params.$\downarrow$\% \\ \hline
\multirow{6}{*}{\begin{tabular}[c]{@{}c@{}}ResNet-56\\ (CIFAR-10)\end{tabular}} & Variational~\cite{zhao2019variational}    & 0.78        & 20.49     \\  
 & KSE~\cite{KSE}      & \textbf{-0.20}   & \textbf{54.73}   \\ 
 & HRank~\cite{lin2020hrank}    & 0.09      & 42.4 \\ 
 & GAL-0.6~\cite{2019Towards}  & 0.28   & 11.8 \\ 
 & L1~\cite{li2016pruning}       & -0.06   & 9.4  \\
 & Ours     & \underline{-0.11}    & 5.4  \\ \hline
\multirow{4}{*}{\begin{tabular}[c]{@{}c@{}}VGG-16\\ (CIFAR-10)\end{tabular}}    & L1~\cite{li2016pruning}     & 0.13     & 64.0      \\ 
 & Variational~\cite{zhao2019variational}      & 0.07   & 73.34 \\ 
 & GAL-0.05~\cite{2019Towards} & 0.19   & \textbf{77.6} \\ 
 & Ours     & \textbf{-0.11}    & 11.6 \\ \hline
\multirow{4}{*}{\begin{tabular}[c]{@{}c@{}}ResNet-56\\ (SVHN)\end{tabular}}    & L1~\cite{li2016pruning}     & -0.01       & 9.1       \\ 
 & SFP~\cite{2018Soft}      & -0.12  & 25.6 \\
 & FPGM~\cite{He_2019_CVPR}     & 0.20    & \textbf{38.8} \\
 & Ours     & \textbf{-0.21}   & 18.7 \\ \hline
\end{tabular}
}
\end{table}

Although our pruning method is inferior to the others in parameter reduction, it is simple yet effective, and the accuracy of ours is competitive. In short, our study shows the great potential in feature disentanglement. This is a pioneering and enlightening research, and  provides a wholly new perspective for network pruning.

\section{Conclusion}

In this paper, we make the first exploration to disentangle the linear features from the output features of a network layer. We devise a simple yet efficient module: the learnable mask module that distinguishes between linear features and non-linear features. We also overcome the non-differentiable problem with the help of the straight-through estimator. Extensive experiments on several datasets demonstrate that our method has a great value in feature disentanglement and provides some insights in the evolution of linear or non-linear features. Besides, our exploration experiments in network pruning have also achieved promising results. Our future work will be on improving the feature disentanglement modules for better parameter efficiency.


\bibliographystyle{IEEEbib}
\bibliography{icmetemplate}

\end{document}